# An Information-Theoretic External Cluster-Validity Measure

Byron E. Dom
IBM Research Division
650 Harry Rd.
San Jose,CA 95120-6099
dom@almaden.ibm.com

## Abstract

In this paper we propose a measure of similarity/association between two partitions of a set of objects. Our motivation is the desire to use the measure to characterize the quality or accuracy of clustering algorithms by somehow comparing the clusters they produce with "ground truth" consisting of classes assigned by manual means or some other means in whose veracity there is confidence. Such measures are referred to as "external". Our measure also allows clusterings with different numbers of clusters to be compared in a quantitative and principled way. Our evaluation scheme quantitatively measures how useful the cluster labels are as predictors of their class labels. It computes the reduction in the number of bits that would be required to encode (compress) the class labels if both the encoder and decoder have free access to the cluster labels. To achieve this encoding the estimated conditional probabilities of the class labels given the cluster labels must also be encoded. In addition to defining the measure we compare it to other commonly used external measures and demonstrate its superiority as judged by certain criteria.

## 1 The Clustering Problem

The most common unsupervised-learning problem is *clustering*, in which we are given a set $\Omega$ of *objects* or *patterns* $\Omega = \{\omega_i | i = 1, 2, \ldots, n\}$ and each object has a representation $x_i \equiv x(\omega_i)$ in some *feature space* which is frequently treated as an $m$-dimensional continuum $\mathbb{R}^m$. Some of the features may be *categorical*, however.

The goal in clustering is to group the objects by grouping their associated *feature vectors* $\mathbf{X} = \{x_i | i = 1, 2, \ldots, n\}$. This grouping can be based on any number of criteria. It is assumed that the dimensions of $x$ are attributes relevant to some application of interest. The grouping is performed on the basis of some measure of similarity relevant to the application and associated feature space. There are numerous objective functions and algorithms for clustering (see [JD88] for a survey), but we are not concerned with these here. Our task is to devise a measure of the quality of the output of clustering algorithms.

Let $K = \{k_i | i = 1, 2, \ldots, n\}$ be a set of cluster labels assigned to the elements of $\mathbf{X}$. The labels themselves are taken from a set $\mathcal{K}$, where $|\mathcal{K}|$ is the number of clusters. We have some clustering procedure $f$ that maps $\mathbf{X}$ to $K$.

**Definition: clustering procedure $f$:**

$$f : \mathbf{X}(\Omega) \rightarrow K(\Omega) \qquad (1)$$

The procedure $f$ may determine the optimal number of clusters as well as the assignment of feature vectors (objects) to class labels or it may accept the number of clusters as input. The set $\Omega$ can be considered to have been drawn from some larger population, which can be characterized by a probability density $p(x)$. The combination of $p(x)$ and the clustering procedure $f$ results in a probability distribution $\{p(k)\}$ over cluster labels.

We define three clustering problems: (1) Each pattern is assigned to one and only one cluster - so-called *partitional* clustering. (2) Each pattern may be assigned to multiple clusters. These are binary assignments. (3) Each pattern has a degree of membership in each cluster. The measure we propose applies to *partitional* clustering. In addition to these three categories a distinction can be made between *flat* and *hierarchical* clustering (although a *flat* is technically a special case of *hierarchical* - i.e. a depth-one tree). Our measure applies to *flat* clustering.

## 2 The Evaluation Problem

*Ex post facto* evaluation of the *validity* (quality or accuracy) of the output of clustering algorithms is a difficult problem in general. Measures or indices of cluster validity can be divided into two types: *external* and *internal*[JD88]. External validity criteria measure how well the clustering results match some prior knowledge about the data. It is assumed that this information is not, in general, computable from $\mathbf{X}$. Perhaps the most common form of external information is a set of classes (categories) and class labels for the objects corresponding to $\mathbf{X}$. These are usually obtained via manual classification.

The use of some measure based solely on the feature data $\mathbf{X}$ (an *internal* measure), begs the question: Why not just use this measure itself as an objective function for clustering? This may in fact be possible in some cases where the objective function used does exactly capture what is desirable in a particular application and there is a feasible algorithm for finding the optimal clustering. In such cases the evaluation problem is moot. In other cases, of course, the answer to this question may have to do with computational feasibility - it may not be possible to devise an algorithm to efficiently find the associated optimal clustering.

In many (if not most) applications, however, clustering algorithms attempt to do what humans can do quite well, albeit slowly relative to the speed of a computer. This human sufficiency is especially true in the case of document clustering, for example, where natural language understanding and vast amounts of world knowledge (or specialized domain knowledge) are used by humans. In such applications the best accuracy/quality measure will therefore be based on human subjective judgments. One way to obtain this is to ask humans to judge the quality of the results directly. This is an expensive and time consuming process however and every algorithm (or variation of a single algorithm) tried will require a new set of subjective judgments.

An alternative to this is to ask humans to cluster the data set into what they consider to be an appropriate set of clusters. This is done *once* to obtain a set $C$ of what we will refer to as *class* labels: $C = \{c_i | i = 1, 2, \ldots, n\}$, $c_i \in \mathcal{C}$. The idea is that the intended users of the algorithm would be quite happy if the algorithm had produced these classes as clusters. They are thus treated as the ideal clustering and quality is judged based on some measure of how well the cluster labels produced by the algorithm(s) agree with the class labels. Any accuracy assessment based on this notion is thus measuring the quality of the clustering *relative* to the particular classification represented by $C$. Another classification will obviously result in a different measure in general. Despite this weakness, *external* measures tend to be the most reliable and are therefore usually preferable when class labels are available In this paper we propose an external validity measure appropriate for flat (non-hierarchical) clustering where a ground-truth classification is available for evaluation purposes.

As in the case of the cluster labels, we can think of $C$ as a sample drawn from a population described by a probability distribution $\{p(c)\}$. Also, we can think of the set of pairs $\{(c_i, k_i)\}$ associated with $C$ and $K$ as a sample drawn from a population described by the distribution $\mathcal{P} \equiv \{p(c, k)\}$.

## 3 Summaries of the Class-Cluster Relationship

A *complete* characterization of the behavior of a particular algorithm when used to cluster a given data set is, of course, contained in the individual objects (e.g. documents) themselves i.e. which objects were assigned to which clusters. Some amount of anecdotal evidence of this type is invaluable in diagnosing the behavior of the clustering algorithm. For large numbers of objects, however, the objects in aggregate are more than can be dealt with in this manner. Some reduced information is essential. For a *partitional* clustering the usual first level of reduction is expressed by the two-dimensional contingency table $\mathcal{H} \equiv \{h(c, k)\}$, where $h(c, k)$ is the number of objects labeled class $c$ that are assigned to cluster $k$ by $f$. In a perfect (from an external measure's point of view) clustering $\mathcal{H}$ is a square matrix (i.e. $|\mathcal{C}| = |\mathcal{K}|$) and only one non-zero element per row/column. Two associated definitions are the one-dimensional *marginal* tables $h(c) \equiv \sum_k h(c, k)$ and $h(k) \equiv \sum_c h(c, k)$.

A further reduction is embodied in the 2 × 2 contingency table $\mathcal{A} = \{a_{ij} | i, j \in \{0, 1\}\}$, where the elements $a_{ij}$ are counts of pairs of vectors $\{x_p, x_q\}$. The row index value $i$ indicates the state of the pairs with respect to the classes. A value of 0 indicates pairs that were assigned to the same class, whereas a value of 1 corresponds to pairs occuring in different classes. Similarly for the column index $j$ except that it corresponds to clusters rather than classes. The symbol "•" in place of an index indicates that the index is summed over (e.g. $a_{0\bullet} = \sum_{i=0}^{1} a_{0i}$.) The table $\mathcal{A}$ is readily computed in terms of $\mathcal{H}$. Formulas are given in [Dom01] and [JD88].

## 4 An External Validity Measure

Our stated intent is to measure the association between $C$ and $K$ by determining how useful (in bits) the latter are in encoding the former. From Shannon[Sha48] we know that the minimum expected "per-symbol" code length for $C$, without the use of $K$, is given by the *entropy* $H(C) = \sum_c p(c) \log p(c)$. On the other hand, the expected per-symbol code length for the class labels given that we know the associated cluster labels and their joint distribution $p(c, k)$ is given by the *conditional entropy*(See [CT91]):

$$H(C|K) = -\sum_{c=1}^{|C|}\sum_{k=1}^{|K|} p(c,k) \log p(c|k).$$

Note that if $k$ is a perfect predictor of $c$, $H(C|K) = 0$, whereas if $k$ has no useful information about $c$, $H(C|K) = H(C)$.

Because we don't know $p(c, k)$, we estimate it using $\mathcal{H}$ and we refer to the associated estimate of $H(C|K)$ as the *empirical* conditional entropy and denote it by:

$$\tilde{H}(C|K) = -\sum_{c=1}^{|C|}\sum_{k=1}^{|K|} \frac{h(c,k)}{n} \log \frac{h(c,k)}{h(k)}. \quad (2)$$

Conditional entropy or its equivalent *mutual information* $(I(C; K) \equiv H(C) - H(C|K))$ has been used by various workers to measure the degree of association between variables. It is incomplete in a certain sense however. To realize the code length given by (2) the decoder must know $\mathcal{H}$ so that it can construct the required decoding tree. Therefore we must have a scheme for encoding $\mathcal{H}$ which will have an associated code length and this code length must be added to (2) to obtain the total code length. Neglecting this second term will usually have a minor effect in comparing two different clusterings as long as they both have the same number of clusters. If these two clusterings have different numbers of clusters, however, this second term (for $\mathcal{H}$) becomes important.

To obtain a code-length for $\mathcal{H}$ we assume an *enumerative* encoding scheme that utilizes the fact that the decoder knows $\{k_i\}$ and therefore $\{h(k)\}$. Think of $\mathcal{H}$ as a matrix with rows indexed by $c$ and columns by $k$. The quantity $h(k)$ is equal to the sum of the elements in the $k^{th}$ column $\mathcal{H}$. The number of possible columns corresponding to $h(k)$ is the number of $|C|$-component vectors with non-negative integer components summing to $h(k)$, which is given by:

$$\binom{h(k)+|C|-1}{|C|-1}. \quad (3)$$

Thus we can encode the $k^{th}$ column by specifying an integer index from 1 through this number, which requires a number of bits equal to the log of (3). This scheme implicitly assumes that all columns consistent with $h(k)$ are equally likely. The number of bits required to encode the entire matrix $\mathcal{H}$ when the $\{h(k)\}$ are already known is thus the log of the number of $\mathcal{H}$'s consistent with $\{h(k)\}$, which is given by:

$$\sum_{k=1}^{|K|} \log \binom{h(k)+|C|-1}{|C|-1}. \quad (4)$$

The use of this encoding scheme can be seen as an application of the *maximum entropy* principle[Jay83] in that this is the most uniform encoding that is consistent with the constraint implied by $\{h(k)\}$. This may be seen more readily by noting this term's Bayesian interpretation wherein two raised to this power (assuming base-two logarithms) is the *prior* probability of observing $\mathcal{H}$ given $\{h(k)\}$.

Our clustering quality measure is the entire encoding cost per object, which is given by:

$$Q_0(C,K) = \tilde{H}(C|K) + \frac{1}{n}\sum_{k=1}^{|K|} \log \binom{h(k)+|C|-1}{|C|-1}. \quad (5)$$

To actually use the encoding scheme implicit in this we would need an additional term of $\log n$ bits to encode $|C|$, but we omit it because it is fixed for a constant for a given ground truth set. This (5) can be seen as an application of the *minimum description length* principle (MDL)[Ris78].

The asymptotic (as $n \to \infty$ for fixed $|C|, |K|$ and $\mathcal{P}$) form of this is derived in [Dom01]. We repeat the result here.

$$Q_0(C,K) \sim H(C|K) + |K|(|C|-1)\frac{\log n}{n}. \quad (6)$$

For very large $n$ the $\frac{\log n}{n}$ term will become insignificant leaving $Q_0(C, K) \sim H(C|K)$.

Our fundamental measure is given by the code length given in (5). As we discuss in the following section, however, most other external measures have the property that the best possible value is equal to one, whereas the worst is zero. Here we transform ours to have this (0, 1] property [1]

$$Q_2(C,K) = \frac{\max_K[Q_0(C,K)] - Q_0(C,K)}{\max_K[Q_0(C,K)] - \min_K[Q_0(C,K)]}. \quad (7)$$

---

[1] In [Dom01] a different form was used for $Q_2$. A user of that form of the measure noted, however, that in certain cases the relative magnitudes of changes in it seemed counterintuitive[Yua02]. While that form still satisfied the desiderata, this form rectifies the observed problem.

Realizing this measure is difficult, however, because of the way in which the value of $\max_K [Q_0(C, K)]$ depends on the details of $\{h(c)\}$. Reasonable results can be obtained using:

$$\min_K [Q_0(C, K)] = \frac{1}{n} \sum_{c=1}^{|\mathcal{C}|} \log \binom{h(c) + |\mathcal{C}| - 1}{|\mathcal{C}| - 1},$$

which is the correct result, and

$$\max_K [Q_0(C, K)] = \tilde{H}(C) + \log |\mathcal{C}|,$$

which is a fairly tight upper bound on the real value of $\max_K [Q_0(C, K)]$.

## 5 Other External Validity Measures

An extensive review of related association measures prior to 1959 including work in the late 19th century can be found in two papers by Goodman and Kruskal[GK54, GK59].

### 5.1 Other use of Information Theoretic Measures

The conditional entropy was used as an external validity measure in [BFR98] and mutual information was used as one in [VD99]. Also workers in the area have discussed mutual information as a measure of association between categorical attributes[JS71, SS73]. We assume that it has not been used more often as an external validity measure for clustering because it is not viable for comparing clusterings with different numbers of clusters.

### 5.2 Classification Error

An external measure that is sometimes used in cases where the number of clusters is equal to the number of classes is *classification error*. If the rows and columns of $\mathcal{H}$ are made to correspond by associating the majority class in each cluster with the cluster itself, then $\mathcal{H}$ can be viewed as the *confusion matrix* of pattern recognition and the sum of its off-diagonal elements divided by the total number of objects is the *total classification error*. The problem with this is that it ignores how incorrect classifications are distributed among the other clusters. Being distributed uniformly randomly among the other clusters is arguably much worse than all going to a single cluster.

Applying this measure to the case where the number of clusters is different than the number of classes is also problematic. This problem can be addressed by defining a "normalized Hamming distance"[HD95] in which the following associations are used:

$$\hat{c}(k) \equiv \arg\max_c h(c, k), \quad (8)$$
$$\hat{k}(c) \equiv \arg\max_k h(c, k). \quad (9)$$

A "directional Hamming distance" $D_H(\cdot; \cdot)$ is defined as

$$D_H(C; K) = \sum_k \sum_{c \neq \hat{c}(k)} h(c, k) \quad (10)$$
$$D_H(K; C) = \sum_c \sum_{k \neq \hat{k}(c)} h(c, k) \quad (11)$$

The *normalized Hamming distance* is then defined in terms of these as follows.

$$\tilde{D}_H(C, K) = 1 - \frac{D_H(C; K) + D_H(K; C)}{2n} \quad (12)$$

### 5.3 Measures Based on $\mathcal{A}$

Jain and Dubes[JD88] list the following four commonly used external measures of partitional validity. All are functions of the matrix $\mathcal{A}$, defined in Section 3.

- Rand[Ran71]: $\frac{a_{00} + a_{11}}{\binom{n}{2}}$

- Jacard[MSS83]: $\frac{a_{00}}{a_{00} + a_{01} + a_{10}}$

- Fowlkes and Mallows[FM83]: $\frac{a_{00}}{\sqrt{\tilde{a}}}$, where $\tilde{a} \equiv a_{0\bullet} a_{\bullet 0}$

- $\Gamma$ statistic(Hubert and Schultz)[HS76]: $\frac{M a_{00} - \tilde{a}}{\sqrt{\tilde{a}(M - a_{0\bullet})(M - a_{\bullet 0})}}$

## 6 Comparison Methodology

How does one compare clustering accuracy measures? By considering only *external* validity measures we have reduced this question to one of what constitutes the best measure of agreement between the two clusterings. Our argument will proceed as follows. While acknowledging that what constitutes a desirable accuracy measure depends on the particular application, we assert that having a general measure is desirable for assessing general clustering algorithms designed to be used in many applications and, even in those cases where a particular application is targeted, measuring the most relevant quantity (e.g. time saved at some particular task) may be infeasible. We believe that our measure is superior to others as a general measure, though we must acknowledge that the choice of which measure to use is, to a certain extent, a matter of taste. We make a concession to this fact in our arguments. First we argue that our measure is superior and should therefore be used in all appropriate cases i.e. comparing a partition to a ground-truth partition. Second, we argue that, even if one doesn't accept that our measure is intrinsically superior, it must at least be acknowledged that it has all the desirable qualitative properties and produces different results from all other measures in some cases and should therefore be accepted as, at least another measure in the set to choose from. In support of these assertions we argue on philosophical grounds that our measure is superior because of its information theoretic basis. We also show that:

(1) our measure has all the desirable properties.

(2) other measures give counter-intuitive (if not simply incorrect) results in certain cases, but that our measure always give the desired behavior.

(3) our measure gives results different from those produced by other commonly used measures.

### 6.1 A Parametric Form for $p(c, k)$

In order to identify desirable properties of a clustering accuracy measure, we define a family of distributions over $(c, k)$ with the hope that this family captures most of the essential characteristics of such distributions from the perspective of characterizing the accuracy of clustering algorithms. Members of this family are identified by values of certain parameters as follows:

- $|\mathcal{C}|$: number of classes

- $(\forall c \in \mathcal{C})\; p(c) = 1/|\mathcal{C}|$

- $|\mathcal{K}|$: number of clusters

- Decomposition of $\mathcal{K}$ into two disjoint subsets $\mathcal{K}_u$ ("useful") and $\mathcal{K}_n$ ("noise"). The cardinalities of these subsets are given by $|\mathcal{K}_u|$ and $|\mathcal{K}_n|$ respectively and clearly $|\mathcal{K}| = |\mathcal{K}_u| + |\mathcal{K}_n|$. The roles of these two cluster subsets are as follows.

    - $\mathcal{K}_n$: These clusters are completely noise:
    
    $(\forall k \in \mathcal{K}_n)\;(\forall c \in \mathcal{C})\; p(c|k) = p(c) = 1/|\mathcal{C}|.$

    - $\mathcal{K}_u$: The clusters in $\mathcal{K}_u$ are correlated with the classes and $\mathcal{K}_u$ is further decomposed into $|\mathcal{C}|$ subsets $\{\mathcal{K}(c)|c \in \mathcal{C}\}$ and correspondingly $\mathcal{C}$ is decomposed into subsets $\{\mathcal{C}(k)|k \in \mathcal{K}_u\}$. The role of these subsets is as follows. For a given $c$ the probabilities $p(k|c)$ are equal for all $k \in \mathcal{K}(c)$. While the sizes of the $\{\mathcal{K}(c)\}$ could be left as parameters also, we determine them automatically as follows.

        * If $|\mathcal{C}| = |\mathcal{K}_u|$, then $|\mathcal{K}(c)| = 1$ and $\mathcal{K}(c)$ consists of cluster $k = c$.

        * If $|\mathcal{C}| < |\mathcal{K}_u|$, then $(\forall k \in \mathcal{K}_u)\; |\mathcal{C}(k)| = 1$ and at least some of the $\{\mathcal{C}(k)\}$ will overlap, corresponding to more than one cluster. Cluster-to-class assignments proceed as follows:

            · The first $\lceil |\mathcal{K}_u|/|\mathcal{C}| \rceil$ clusters are assigned to class 1.
            · The next $\lceil (|\mathcal{K}_u|-|\mathcal{C}(1)|)/(|\mathcal{C}|-1) \rceil$ clusters are assigned to class 2.
            · and so on .... The number of clusters to be assigned to the next class being given by the *ceiling* of the ratio of the number of unassigned clusters remaining to the number of unassigned classes remaining.

        * If $|\mathcal{C}| > |\mathcal{K}_u|$, then the class-to-cluster assignments proceed in a manner exactly analogous to the cluster-to-class assignments in the $|\mathcal{C}| < |\mathcal{K}_u|$ case.

- $\epsilon$: total error probability:

$$\sum_{c \in \mathcal{C}} \sum_{k \notin \mathcal{K}(c)} p(c,k) = \epsilon$$

$$(\forall c \in \mathcal{C})\;(\forall k \in \mathcal{K}(c))\; p(k|c) = \frac{1-\epsilon}{|\mathcal{K}(c)|}$$

- $\epsilon_1, \epsilon_2$: error components: $\epsilon = \epsilon_1 + \epsilon_2$

$$(\forall c \in \mathcal{C})\;[\forall (k \in \mathcal{K}_u) \wedge (k \notin \mathcal{K}(c))]$$
$$p(k|c) = \frac{\epsilon_1}{|\mathcal{K}_u| - |\mathcal{K}(c)|}$$

$$(\forall c \in \mathcal{C})\;(\forall k \in \mathcal{K}_n)\; p(k|c) = \frac{\epsilon_2}{|\mathcal{K}_n|}$$

$$\sum_{c \in \mathcal{C}} \sum_{k \in \mathcal{K}_n} p(c,k) = \epsilon_2$$

An example $\{p(c, k)\}$ is presented in Table 1.

Table 1: Class-cluster joint probability distribution $\mathcal{P}$ with $|\mathcal{K}_u| = |\mathcal{C}| = 5$, $|\mathcal{K}_n| = 3$, $\epsilon_1 = 0.2$ and $\epsilon_2 = 0.3$. clusters $\{6, 7, 8\}$ are *noise*, while $\{1, 2, 3, 4, 5\}$ are *useful*.

| class ↓ | ← cluster → | | | | | | | |
|---|---|---|---|---|---|---|---|---|
| | 1 | 2 | 3 | 4 | 5 | 6 | 7 | 8 |
| 1 | 0.10 | 0.01 | 0.01 | 0.01 | 0.01 | 0.02 | 0.02 | 0.02 |
| 2 | 0.01 | 0.10 | 0.01 | 0.01 | 0.01 | 0.02 | 0.02 | 0.02 |
| 3 | 0.01 | 0.01 | 0.10 | 0.01 | 0.01 | 0.02 | 0.02 | 0.02 |
| 4 | 0.01 | 0.01 | 0.01 | 0.10 | 0.01 | 0.02 | 0.02 | 0.02 |
| 5 | 0.01 | 0.01 | 0.01 | 0.01 | 0.10 | 0.02 | 0.02 | 0.02 |

### 6.2 Desirable Characteristics of Clustering Accuracy Measures

Assume that some accuracy measure $\mathcal{M}(\mathcal{H})$ increases monotonically from 0 to 1 as accuracy improves. Assume further that we intend to apply it to the expected $\mathcal{H}$ or $\mathcal{A}$ corresponding to the family of distributions just defined above. For a fixed $n$ and $|\mathcal{C}|$ what behavior do we desire of $\mathcal{M}$ with respect to the parameters of this family: $|\mathcal{K}_u|, |\mathcal{K}_n|, \epsilon_1$ and $\epsilon_2$? The answer, expressed in terms of differences and derivatives with respect to single parameters while all others are held fixed, is as follows:

**P1:** $\mathcal{M}$ should decrease (indicating a worse clustering) whenever the number $|\mathcal{K}_u|$ of *useful* clusters varies away from the number $|\mathcal{C}|$ of classes.

  **P1.1:** For $|\mathcal{K}_u| < |\mathcal{C}|$ and $\Delta|\mathcal{K}_u| \leq (|\mathcal{C}| - |\mathcal{K}_u|)$, $\frac{\Delta \mathcal{M}}{\Delta|\mathcal{K}_u|} > 0$.

  **P1.2:** For $|\mathcal{K}_u| \geq |\mathcal{C}|$, $\frac{\Delta \mathcal{M}}{\Delta|\mathcal{K}_u|} < 0$.

**P2:** $\mathcal{M}$ should decrease (indicating a worse clustering) whenever the number of *noise* clusters $|\mathcal{K}_n|$ increases That is, we should have $\frac{\Delta \mathcal{M}}{\Delta|\mathcal{K}_n|} < 0$.

**P3:** $\mathcal{M}$ should decrease (indicating a worse clustering) whenever either of the error parameters $\epsilon_1$ and $\epsilon_2$ increases:

  **P3.1:** $\frac{\partial \mathcal{M}}{\partial \epsilon_1} \leq 0$ with equality holding only when $|\mathcal{K}_u| = 1$.

  **P3.2:** $\frac{\partial \mathcal{M}}{\partial \epsilon_2} \leq 0$ with equality holding only when $|\mathcal{K}_n| = 0$.

### 6.3 Analysis of $Q_{0,2}$ Vis a Vis the Desiderata

For simplicity we will base our analysis/discussion on the asymptotic forms of our measure given in (6). Here we consider the variation with respect to the various model parameters.

- **P1.1 and P1.2** ($|\mathcal{K}_u|$): Certainly, increasing $|\mathcal{K}_u|$ beyond $|\mathcal{C}|$ will increase both $H(C|K)$ and the model cost. Decreasing it from $|\mathcal{C}|$ will increase $H(C|K)$, while decreasing the model cost, however. We see from (6) that the increase in $H(C|K)$ will dominate, asymptotically.

- **P2** ($|\mathcal{K}_n|$): Increasing $|\mathcal{K}_n|$ while holding $\epsilon_2$ fixed will increase the model cost while having no effect on $H(C|K)$. The former is obvious, while the latter is due to the fact that only the fraction $\epsilon_2$ of objects that are assigned to noise clusters (not the number of noise clusters) affects $H(C|K)$.

- **P3.1 and P3.2** ($\epsilon_1$ and $\epsilon_2$): Increasing either $\epsilon_1$ or $\epsilon_2$ will clearly increase $H(C|K)$, but have no effect on the asymptotic model cost $|\mathcal{K}|(|\mathcal{C}| - 1)\log n$. Thus $\frac{\partial Q_0}{\partial \epsilon_{1,2}} \geq 0$, which implies that $\frac{\partial Q_2}{\partial \epsilon_{1,2}} \leq 0$.

Thus we have shown that our measure satisfies all of our desired characteristics asymptotically. In the following section we will do an exploration of the ability of all of the measures discussed here to satisfy these characteristics for a certain set of test cases.

## 7 Comparison Results

Our comparison of the measures proceeds as follows. All results are obtained by first computing the expected values of $\mathcal{H}$ and $\mathcal{A}$ for the given $\mathcal{P}$ (i.e. corresponding to specific values of the parameters $|\mathcal{C}|, |\mathcal{K}_u|, |\mathcal{K}_n|, \epsilon_1$ and $\epsilon_2$) and then using them to compute values of the various validity measures In most of the results we present the $Q_2$ form of our measure because it has the same general behavior as the other measures in that $Q_2 = 1$ corresponds to a perfect clustering and values decrease toward 0 as the clustering accuracy gets worse. Both forms of the measure ($Q_0$ and $Q_2$) obviously produce the same ordering of a given group of clusterings all compared against the same ground truth. Due to space limitations a subset of the characterization results is presented here. The complete results can be found in [Dom01].

We begin by presenting a simple case (See Figure 1), which has $|\mathcal{C}| = |\mathcal{K}| = |\mathcal{K}_u| = 5$ and $|\mathcal{K}_n| = 0$. The $\epsilon_1$ parameter is then varied over a range of $[0, 0.8]$. All measures show the desired (and expected) behavior with their values decreasing with increasing $\epsilon_1$. Most plots of the behavior of the measures while any one of the error parameters is varied while holding the others fixed have a similar qualitative character.

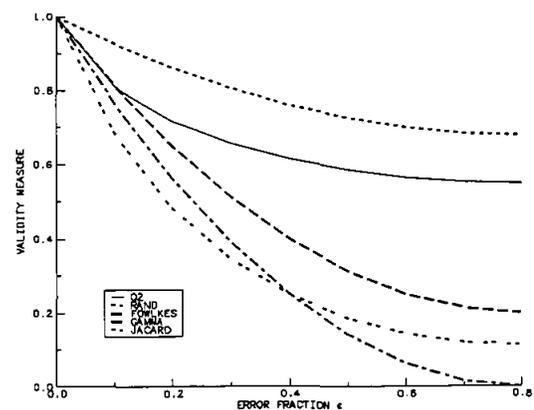

Figure 1: The result of computing several external validity measures including $Q_2$ while varying the error parameter $\epsilon_1$ and holding $|\mathcal{C}| = |\mathcal{K}| = |\mathcal{K}_u| = 5$ ($|\mathcal{K}_n| = 0$).

To determine if all measures satisfy the desiderata the following detailed comparison over a broad range of model parameters was run. The calculated accuracy-measure values reported in this section correspond to fixed values of $|\mathcal{C}|(|\mathcal{C}| = 5)$ and $n(n = 500)$. All other model parameters were varied over the following values.

- $|\mathcal{K}_u|$: 10 values: $2, 3, \ldots, 11$

- $|\mathcal{K}_n|$: 7 values: $0, 1, \ldots, 6$

- $\epsilon_1$: 4 values: $0, 0.066\ldots, 0.133\ldots, 0.2$

- $\epsilon_2$: 4 values: $0, 0.1, 0.2, 0.3$

Running through the complete range of these values would result in $1120 = 10 \times 7 \times 4 \times 4$ instances of $\{p(c,k)\}$, but when meaningless combinations (e.g. $|\mathcal{K}_n| = 0$ with non-zero $\epsilon_2$ values) are eliminated, there are 760 valid combinations. For each valid parameter combination values of all 7 measures ($Q_0$, $Q_2$, Rand, Fowlkes, Gamma, Jacard and Hamming) were calculated. Two measures mentioned in Section 5, $\tilde{H}(C|K)$ and classification error, are omitted in this comparison because, as discussed in previous sections, they are not suitable for comparing clusterings with different numbers of clusters. This problem is repaired for classification error by $\tilde{D}_H(C,K)$ and for $H(C|K)$ by $Q_0(C,K)$.

While this is clearly not a complete characterization of the behavior of these measures, it has the following objectives.
(1) To confirm the analysis of our measure in Section 6.3. The absence of any violations of our desired characteristics, while not proving that they hold in all cases, is evidence in support of that contention.
(2) To show that our measure produces a unique ranking of the various cases, a ranking different from that produced by other measures.
(3) To see if any of the other measures violate our desirability criteria in any of these cases, which would suggest that our measure is superior as judged by our criteria.

The results of these calculations and their associated ranks according to all the measures are presented in [Dom01]. The rank results for one measure (Fowlkes and Mallows[FM83] are plotted in Figure 2. The rank results and measure values for all the measures appear qualitatively similar to this figure. They are omitted here due to space limitations. In Figure 2 the F&M ranks for the various $\{p(c,k)\}$ models are plotted versus those produced by our measure. As can be seen from these figures, the values of the measures and their rankings over the range of model-parameter values explored, the other measures, while clearly correlated with our measure ($Q_2$), produce different rankings. Thus our measure is a clear alternative with its own unique ranking.

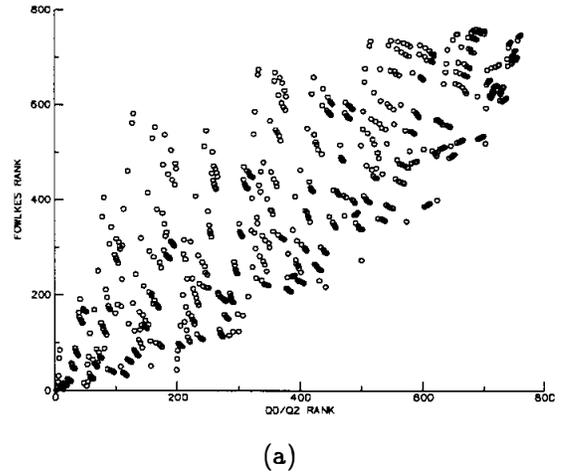

(a)

Figure 2: A comparison of the ranks produced by the Fowlkes-and-Mallows measure and $Q_2$

### 7.0.1 Examination of Whether or not the Measures Satisfy the Desirable Properties of Section 6.2

We analyzed the results of our calculations to see if the desirable properties listed in Section 6.2 were satisfied. Checking all instances of the associated differences, we found the following.

- Our measure ($Q_2$) satisfied all tests.

- **P3.1**: All measures satisfied all tests of the $\frac{\partial \mathcal{M}}{\partial \epsilon_1} < 0$ criterion.

- **P3.2**: All measures except Rand satisfied all tests of the $\frac{\partial \mathcal{M}}{\partial \epsilon_2} < 0$ criterion. We observed 29 cases where Rand failed this test. All occured for cases where $|\mathcal{K}_u| < |\mathcal{C}|$; most for $|\mathcal{K}_u| = 2$ and only 4 for $|\mathcal{K}_u| = 3$ with none for $|\mathcal{K}_u| \geq 4$.

- **P1.1** and **P1.2**: We observed 14 instances of measures failing the $\frac{\Delta \mathcal{M}}{\Delta |\mathcal{K}_u|}$ tests. Of these 12 were Rand and 2 were Hamming. The two Hamming cases were minor infractions where $\frac{\Delta \mathcal{M}}{\Delta |\mathcal{K}_u|} = 0$ was observed between $|\mathcal{K}_u| = 10$ and $11$ for particular instances of the other model parameters. All the Rand errors involved detecting the peak at either $|\mathcal{K}_u| = 6$ or $7$. All were for large values of the error parameters $\epsilon_1$ and $\epsilon_2$.

- **P2**: The test that created the most problems for the measures other than ours was $\frac{\Delta \mathcal{M}}{\Delta |\mathcal{K}_n|} < 0$. This difference ratio was measured for each of the measures at each of 6 $|\mathcal{K}_n|$ values $1, 2, \ldots, 6$ for every combination of the following values of the other three model parameters.

  - $|\mathcal{K}_u|$: 10 values: $2, 3, \ldots, 11$

- $\epsilon_1$: 4 values: $0, 0.066..., 0.133..., 0.2$
- $\epsilon_2$: 3 values: $0.1, 0.2, 0.3$

Any time one or more violations of this criterion were detected in a sequence of measure values corresponding to the sequence of $|\mathcal{K}_n|$ values $1, 2, \ldots, 6$, it was counted as one violation. The distribution of these violations over the measures is given in Table 2. Note that three of the measures (Rand, Gamma and Hamming) failed on every one of the 120 instances of the parameter triple $(|\mathcal{K}_u|, \epsilon_1, \epsilon_2)$. The Hamming measure failed because it showed no sensitivity to this parameter i.e. $\frac{\Delta \mathcal{M}}{\Delta |\mathcal{K}_n|} = 0$.

Table 2: The number of cases (out of 120) where the various measures failed the $\frac{\Delta \mathcal{M}}{\Delta |\mathcal{K}_n|} < 0$ test.

| Rand | Fowlkes | Gamma | Jacard | Hamming |
|------|---------|-------|--------|---------|
| 120  | 103     | 120   | 80     | 120     |

## 8 Discussion and Conclusion

We have proposed and evaluated a new external cluster-validity measure based on information-theoretic considerations. We have also examined the behavior of this measure and its ability to satisfy certain desirability criteria. We have also compared it with other commonly used external validity measures.

The answer to the question of which clustering accuracy measure is *best* will depend on the particular application and it is certainly impossible to anticipate every possible application. In those cases where it is judged that accuracy is best measured by comparing the results of a clustering algorithm to some ideal ("ground truth"), however, our measure is appropriate. At the very least, we can say that the measure we propose offers one more choice to the list of similarity measures appropriate for such comparisons and this measure may give a relative ranking (among various clustering results) that is different from that produced by other measures. We would like to say something stronger, however. We believe that the measure proposed here is superior to other measures in a certain fundamental sense. *Information theory* has, since its inception in 1948[Sha48], clearly demonstrated the viability of code length as a measure of information content. The subsequent development of the theory of *algorithmic complexity*[Kol65] extended these ideas and ultimately led to the *minimum description length principle*[Ris78], which distilled the essence of these and extended them further. For this reason we feel that our measure, which embodies these principles, is superior to the other measures discussed here, which we consider to be more heuristic in nature. This is, of course, a philosophical argument. In support of it we have shown that, when compared with other measures, our measure is the only one that satisfies all of a set of desiderata related to how measure values should vary with certain features of the class-cluster distribution.

## Acknowledgment

We would like to thank Alex Cozzi for many helpful and detailed comments after a careful reading of an earlier draft of this document. We also thank Jun Yuan for the experimental feedback and related discussion that resulted in changing the definition of $Q_2$. Finally we thank the three anonymous reviewers for their helpful comments, though circumstances force us to defer addressing some of them until a later date.